\documentclass[12pt,journal]{article}
\usepackage{times,latexsym}
\usepackage{url}
\usepackage[T1]{fontenc}

\usepackage{amsmath}
\usepackage{amsfonts}
\usepackage{amssymb}
\usepackage{natbib}
\usepackage{algorithm}
\usepackage{algpseudocode}
\usepackage{amsthm}
\usepackage{dsfont}
\usepackage{tabularray}
\usepackage{bbm}
\usepackage{dsfont}
\usepackage{graphicx}
\usepackage{mathtools}
\usepackage[toc,page]{appendix}
\usepackage{float}

\numberwithin{equation}{section}
\DeclareMathOperator*{\argmax}{arg\,max}

\begin{document}

\title{
Semantic Tagging with LSTM-CRF
}
\author{Farshad Noravesh\footnote{Email: noraveshfarshad@gmail.com}}
\maketitle


\begin{abstract}
In the present paper, two models are presented namely LSTM-CRF and BERT-LSTM-CRF  for semantic tagging of universal semantic tag dataset. The experiments show that the first model is much easier to converge while the second model that leverages BERT embedding, takes a long time to converge and needs a big dataset for semtagging to be effective.
\end{abstract}

\section{Introduction}
Tagging can always be seen as an initial step in any task such as dependency parsing as is done in \citep{Vacareanu2020} or part of speech(POS) tagging as well as named entity recognition(NER) tagging.  

POS tagging as well as NER tagging for semantic parsing is very restricted and they determine lexical semantics with some shortcomings. Univeral semantic tagging(semtagging) is motivated to reduce and compensate such limitations and shortcomings. 
Another motivation is that parsing community are shifting from syntactic dependency tree parsing to semantic dependency graph parsing and semtagging could be seen as an initial step in these investigations.

Semantic tagging is the task of assigning language-neutral semantic categories to words. The necessity of semantic tagging  can be well realized in recent research on semantic parsing. \citep{Zheng2020} decomposes semantic parsing into two parts. In the first part, input utterance $x$ is tagged with semantic symbols. In the second part, a sequence to sequence model is used to use these semantic features to produce the final semantic parsing which could be represented in different meaning formalisms like lambda calculus or SQL queries. The semantic labels $z$ in \citep{Zheng2020} are unobserved and is considered as a latent variable which is learned to represent $p(z|x;\theta)$ where $\theta$ are the parameters of the model that is learned by a BiLSTM model.

 \citep{Bjerva2016} uses deep residual networks, and enters the inputs as both word and character representations for the task of semantic tagging. They used the Groningen Meaning Bank (GMB) corpus as well as the Parallel Meaning Bank (PMB) datasets and obtained better signal propagation as well as less overfitting in these deep networks.

Semantic tagging has two major applications. One can use it for multitask learning like \citep{Abdou2018} or using it in a pipeline to improve quality of vector representations for downstream tasks such as machine translation as described in \citep{Belinkov2018}.

\section{Modeling Semantic Tagging}
\citep{Huang2015} uses a combination of biLSTM and CRF to predict tagging problem. \citep{Lample2016} improves the biLSTM-CRF model by better vector representation of the words which considers both compositional form and function which is inspired by \citep{Ling2015}.  \citep{Ma2015} combines biLSTM-CRF with convolutional neural networks.
\begin{figure} [h!]
\centering
  \includegraphics[width=100mm,scale=0.5]{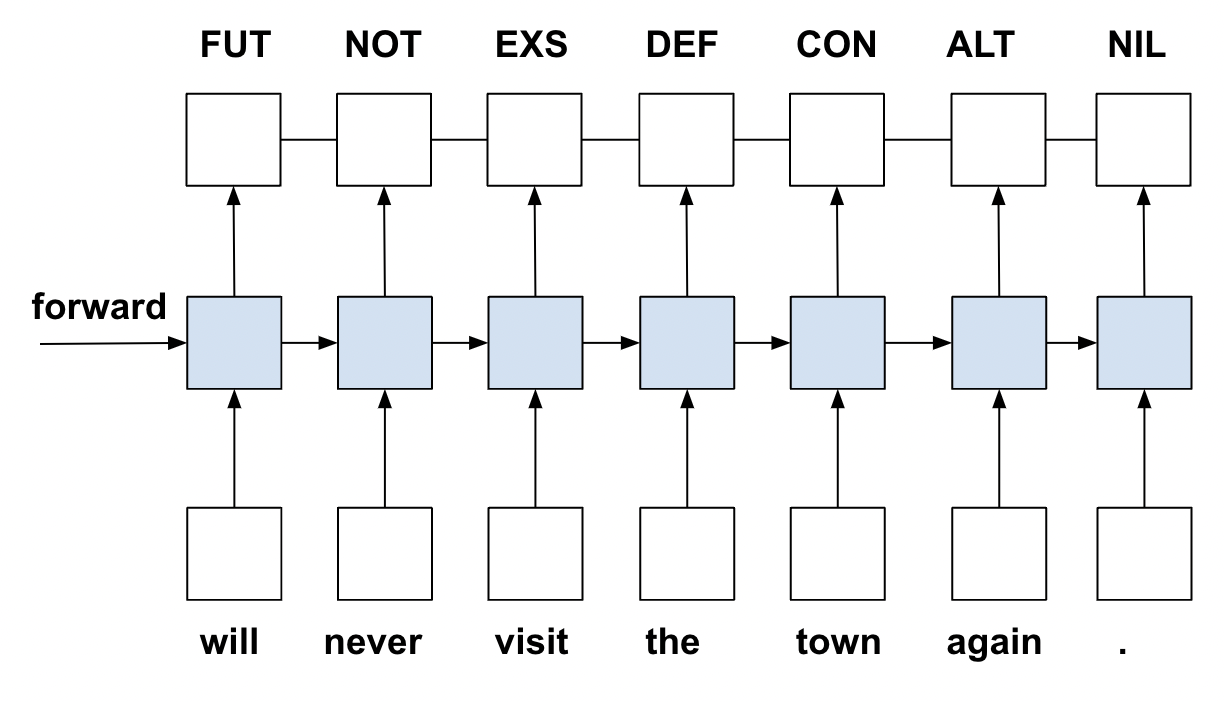}
  \caption{LSTM-CRF model for semtagging}
  \label{fig:model} 
\end{figure}

\subsection{LSTM-CRF Model}
Semtagging dataset in \citep{Abzianidze2017} has been used for training in the present paper. 73 sem-tags are grouped into 13 meta-tags. The baseline in the present paper is similar to \citep{Huang2015} but LSTM is used instead of biLSTM , and \citep{Huang2015} used it for named entity recognition while the present paper focuses on semantic tagging prediction. Figure~\ref{fig:model} shows the architecture of LSTM-CRF that takes a sentence as a sequence of words $x=x_{1},\ldots,x_{T}$ and outputs a sequence of semantic tags $y=y_{1},\ldots,y_{T}$. 
Instead of using pretrained word embedding like GloVe, LSTM-CRF model in the present paper learns the embedding in an End-To-End approach by modeling the embedding as a simple linear layer that maps token indices to word vector representation and the weights of this linear model is learned along with the weights of LSTM and CRF. 
The features are produced by the LSTM and CRF use tag informations and learns the parameters of transition matrix. The matrix of the scores which is output by the model are denoted by $f$ and its entries are $[f_{\theta}]_{t,y_{t}}$ which refers to the score of semtag $y_{t}$ at word in time $t$ with model parameters $\theta$ and is called emission score. 
$A$ is the matrix of transition scores which is position independent(shared across time steps) and its entries $[A]_{y_{t},y_{t+1}}$ denotes the transition score from state $y_{t}$ to state $y_{t+1}$ for a pair of consecutive time steps . The score for a path of tags $Y$ for a sentence $X$ is given by the sum of transition scores and emission scores. The goal is to learn matrix $A$ and parameters $\theta$
\begin{equation}
s(x,y)=\sum_{t=1}^{T}( [A]_{y_{t},y_{t+1}} +[f_\theta]_{t,y_{t}} )
\end{equation}
Now, the probability of a semtag sequence y is the following softmax:
\begin{equation}\label{cond}
p(y|x)=\frac{e^{s(x,y)}}{\sum_{\tilde{y}\in Y_{X}} e^{s(x,\tilde{y})}}
\end{equation}
where $Y_{X}$ in \eqref{cond} represents all possible semtag sequences. The following log-probability of gold semtag sequence should be maximized during training.
\begin{equation}\label{final_cond}
\log p(y|x)= s(x,y) -\log \sum_{\tilde{y}\in Y_{X}} e^{s(x,\tilde{y})}
\end{equation}
The second term in \eqref{final_cond} is the logarithm of partition function and can be calculated efficiently using forward algorithm($\alpha$-algorithm) which is a dynamic programming algorithm.
Once the transition matrix parameters and emission function parameters are learned, Viterbi algorithm is used to obtain the most likely path of semtag sequence. Viterbi variable can be obtained using dynamic programming which is a datastructure that has the following recursive relation: 
\begin{equation} \label{viterbi}
\pi[j,s]= \underset{s'\in 1,\ldots,k}\max \pi[j-1,s']\times A(s|s') \times f(x_{j}|s)
\end{equation}
Viterbi algorithm is also used during training.  $\pi[j,s]$ in \eqref{viterbi} is the maximum probability of a sequence ending in state $s$ at time $j$. Backpointers $bp[j,s]$ are also recorded during Viterbi algorithm:
\begin{equation} \label{backpointers}
bp[j,s]= \underset{s'\in 1,\ldots,k}\argmax \pi[j-1,s']\times A(s|s') \times f(x_{j}|s)
\end{equation}

\subsection{BERT-LSTM-CRF Model}
\begin{figure} [h!]
\centering
  \includegraphics[width=100mm,scale=0.5]{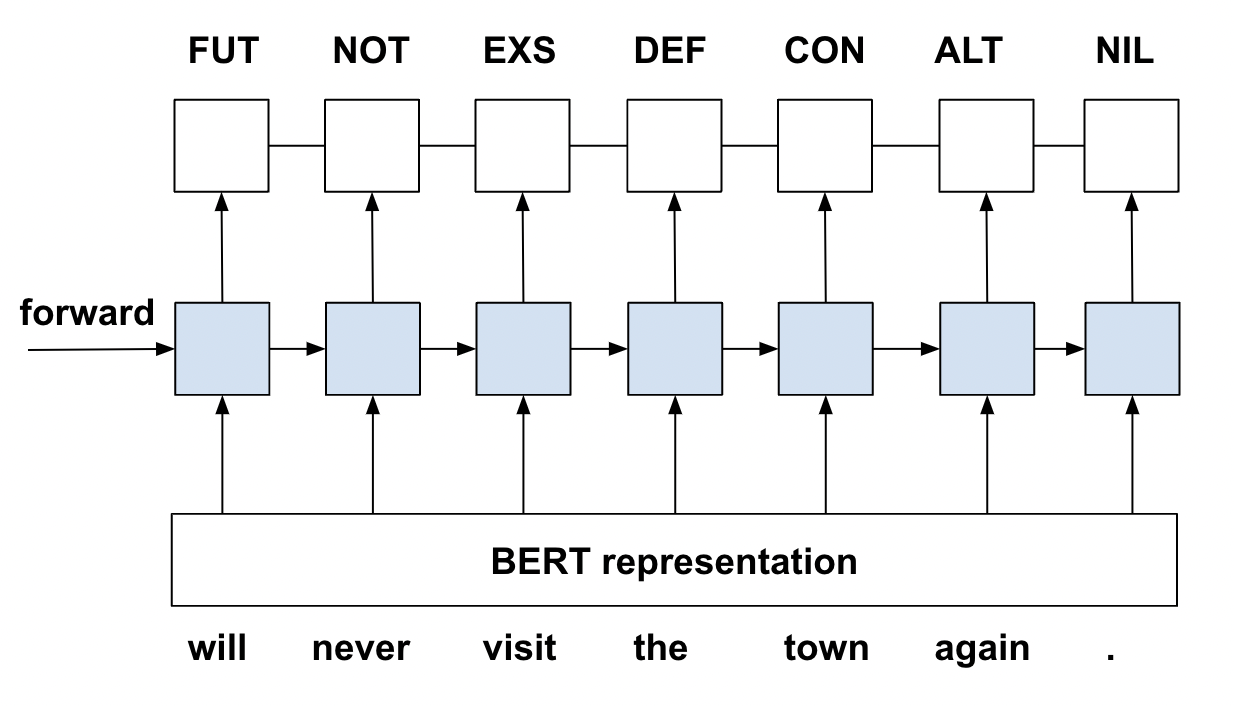}
  \caption{BERT-LSTM-CRF model for semtagging}
  \label{fig:secondModel} 
\end{figure}

Another approach for word representation is BERT  \citep{Devlin2019}.
The second model in the present paper uses BERT for dynamic embedding of words.
BERT embeddings captures contextual information better than pre-trained embeddings or traditional embedding as is noticed in \citep{He2019}. Thus, BERT embedding is investigated as a vector representation for the input given to LSTM-CRF semtagging model.
Context-informed word embeddings like BERT capture other forms of information that result in more accurate feature representations than traditional word2vec algorithms. Word2Vec has a fixed representation and global while BERT has a dynamic representation which is conditioned on the context inside the given sentence and is very subjective. The second model in the present paper is shown in Figure~\ref{fig:secondModel} which is called BERT-LSTM-CRF. Since the dataset for semtagging is small, the second model which suffers from curse of dimensionality needs more data to tune the weights of the model.

\subsection{Experiments}
All experiments parameters are listed in table~\ref{fig:experiments}. The last experiment is done using BERT-LSTM-CRF while the rest of them are based on LSTM-CRF model. The number of epochs in all these experiments is fixed to 20 to be able to easily realize and observe the rate of convergence at different settings. Appendix~\ref{appendix:res} shows the training and validation accuracy and loss for all experiments. A dynamic learning rate has been used  in a way that is reduced automatically every 10 epoch by a factor of 0.1 in all experiments.
\begin{table}[h!]
  \begin{center}
    \caption{different experiments for semtagging}
    \label{fig:experiments}
	\begin{tblr}{h{3em}h{3em}h{3em}h{3em}h{3em}h{3em}}
	\hline
	{ex} & {opt} & {epochs}& {batch size} &{embDim} &{hidDim} \\
      	\hline
      	1 & Adam & 20 & 5 & 50 & 8 \\
      	2 & Adam & 20 & 5 & 100 & 20 \\
	3 & SGD & 20 & 5 & 100 & 20 \\
	4 & Adam & 20 & 20 & 100 & 20 \\
	5 & Adam & 20 & 5 & 100 & 30 \\
	6 & Adam & 20 & 5 & 100 & 50 \\
	7 & SGD & 20 & 5 & 768 & 600 \\
    \end{tblr}
  \end{center}
\end{table}

\begin{figure}  
\centering
  \includegraphics[width=100mm,scale=0.5]{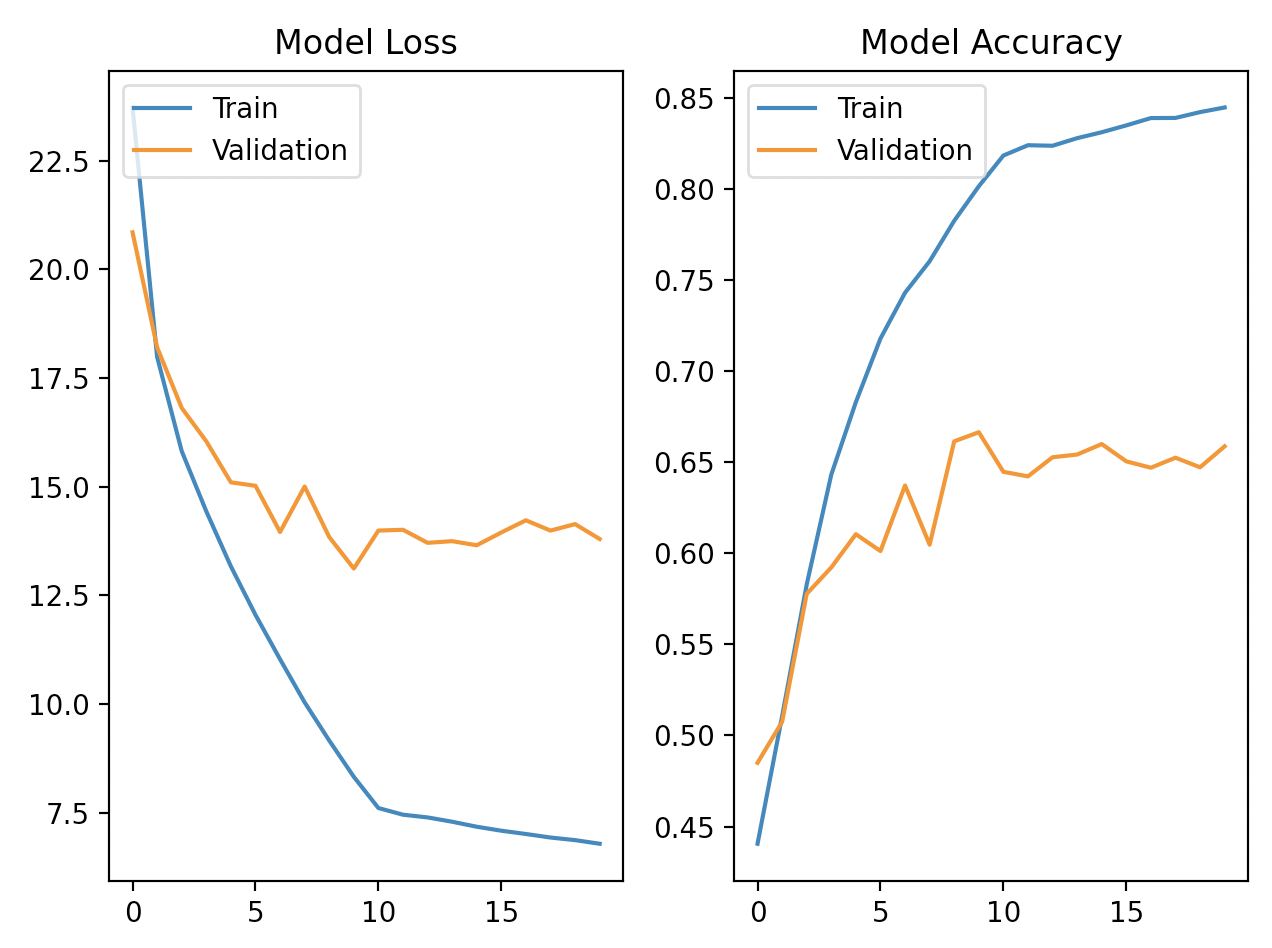}
  \caption{experiment 7}
  \label{fig:ex7} 
\end{figure}

Experiment 6 shows a training accuracy of 95 percent and validation accuracy of 89 percent since model complexity is increased by setting embedding dimension of 100 and hidden dimension of 50 as is shown in Table~\ref{fig:experiments}. 
Experiment 7 in table~\ref{fig:experiments} uses BERT embedding instead of an internal embedding layer and is harder to converge for the following main reason. The reason is because the hidden size is 768 for word embedding and this creates the curse of dimensionality since there is not enough data for semantic tagging and this added complexity makes the overall model very data hungry.
The fluctuations in Figure~\ref{fig:ex7} are for three reasons. The first reason originates from higher model complexity that hidden dimension of BERT creates. The second reason is for using relatively smaller batch size. The third reason which is the least significant factor arises from using SGD optimizer instead of Adam optimizer.

 \section{Conclusion}
 The importance of semtagging and some applications of it are explained. It is shown how LSTM-CRF and  BERT-LSTM-CRF can predict semantic tags and the first model converges quickly even with small dataset since model complexity is relatively low. 
 It should be emphasized that semtagging could have major impact on semantic parsing improvement in all formalisms such as lambda calculus, abstract meaning representation (AMR), discourse representation structure(DRS).
A research direction is improving semantic operator prediction in \citep{Noravesh2023} either by augmenting POS tags with semtags or using semtags and pretrained word embedding. Another research direction is using knowledge distillation to have a low complex model since the current dataset for semtagging is relatively small for word embedding using BERT which has the default size of 768. Many knowledge distillation models have been done in the literature like \citep{Sanh2019} which produce models with smaller complexity which is suitable for small datasets like universal semantic tagging dataset.

\section{Appendix}
\label{appendix:res}
\begin{figure} [!h]
\centering
  \includegraphics[width=100mm,scale=0.5]{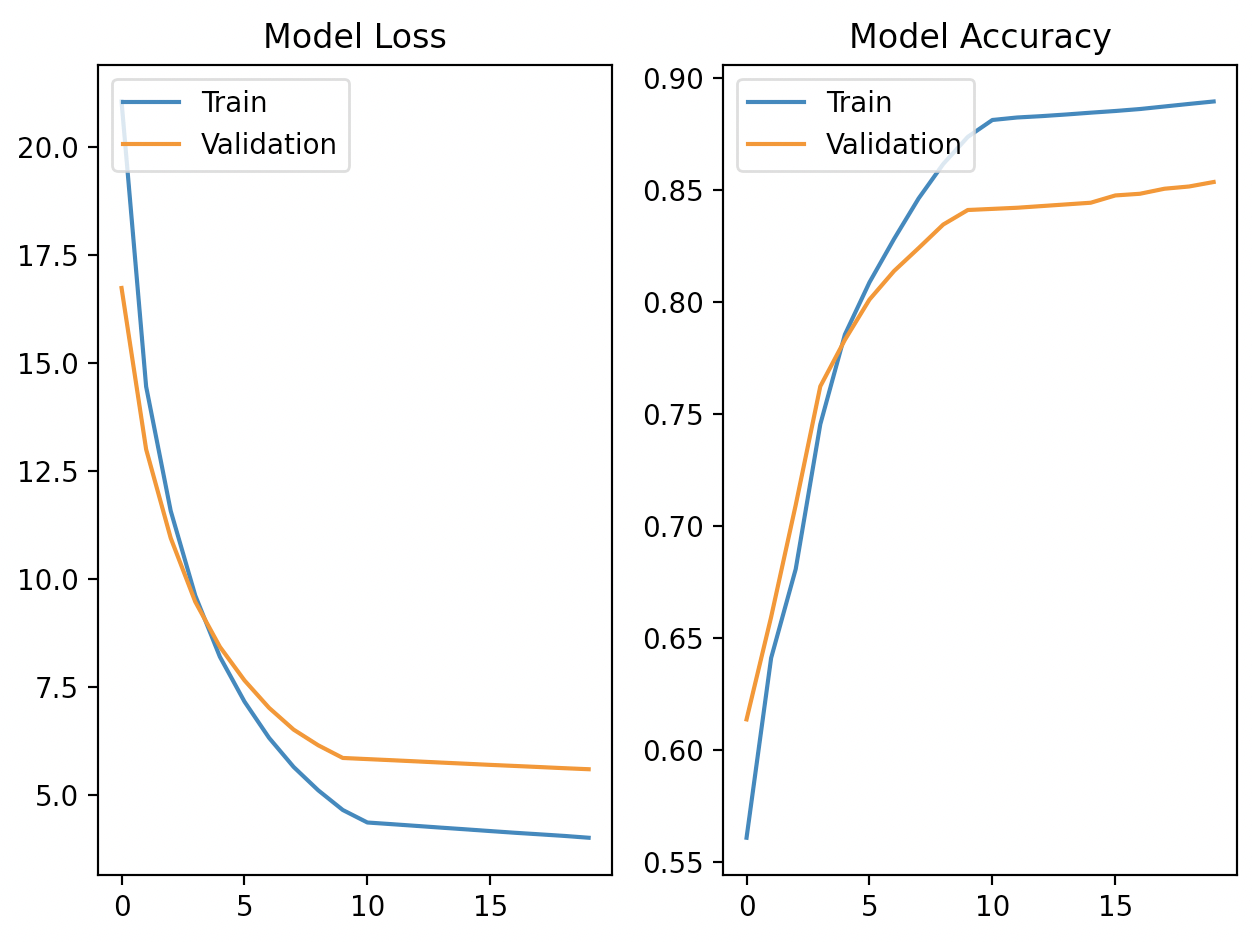}
  \caption{experiment 1}
  \label{fig:ex1} 
\end{figure}

\begin{figure} 
\centering
  \includegraphics[width=100mm,scale=0.5]{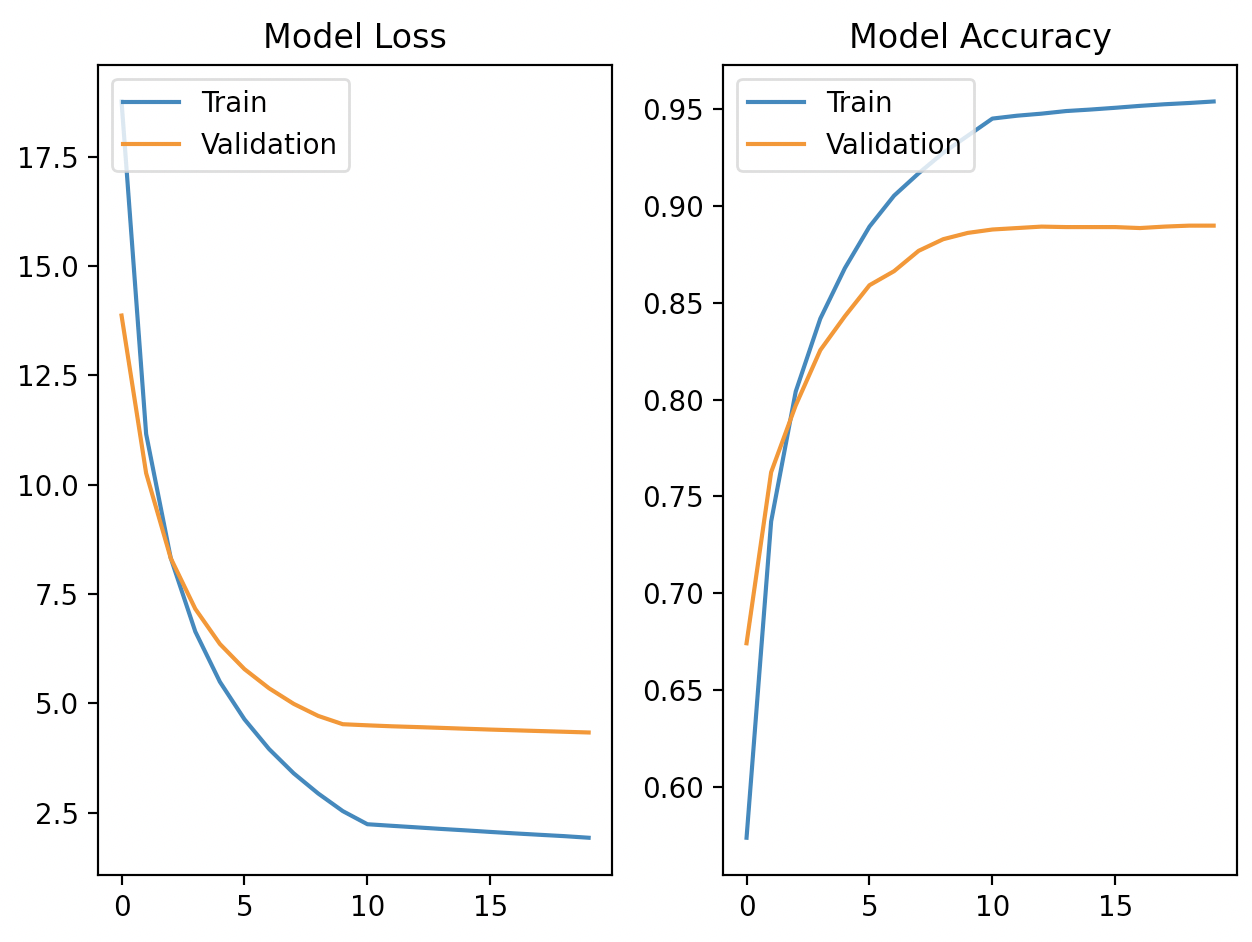}
  \caption{experiment 2}
  \label{fig:ex2} 
\end{figure}

\begin{figure} 
\centering
  \includegraphics[width=100mm,scale=0.5]{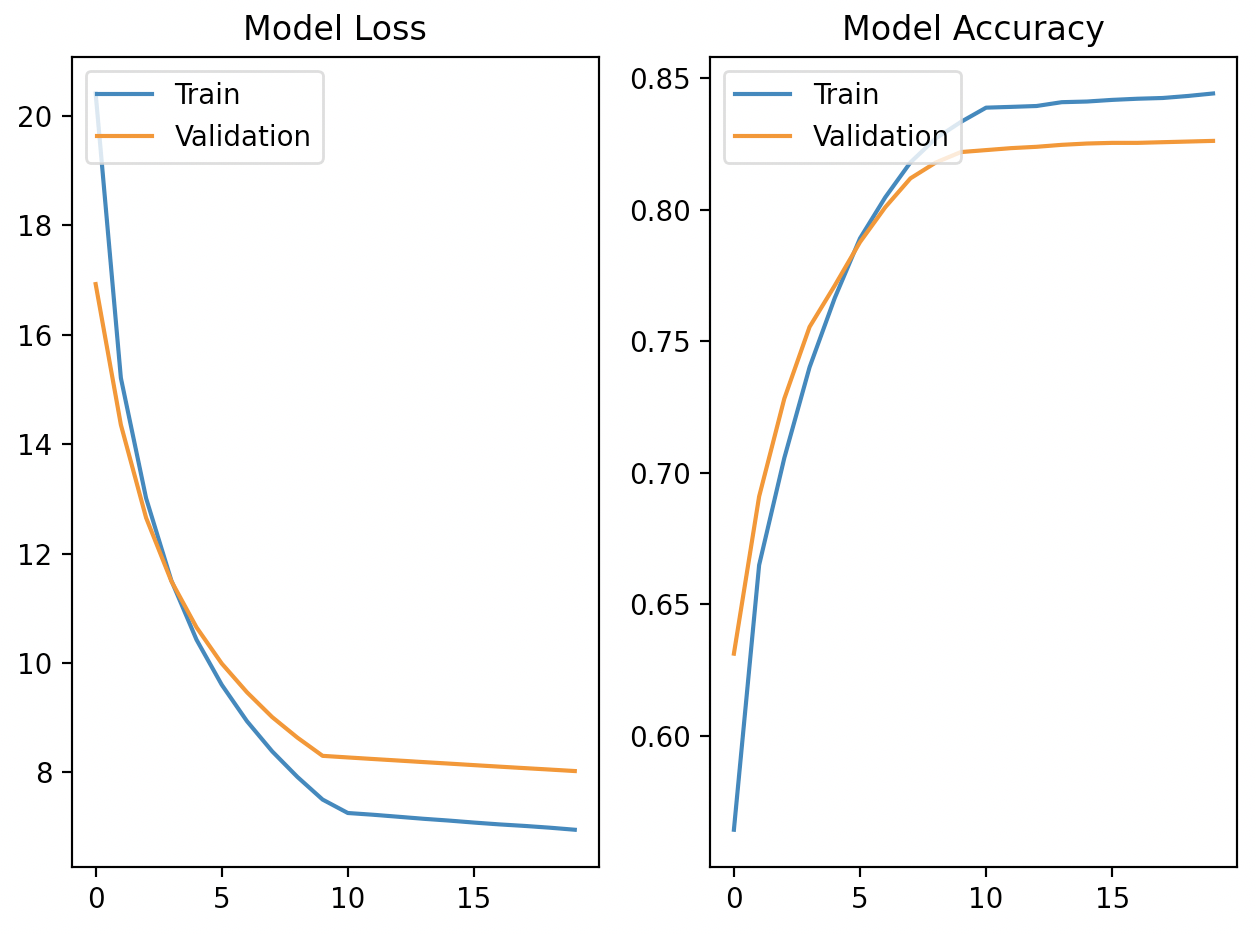}
  \caption{experiment 3}
  \label{fig:ex3} 
\end{figure}

\begin{figure} 
\centering
  \includegraphics[width=100mm,scale=0.5]{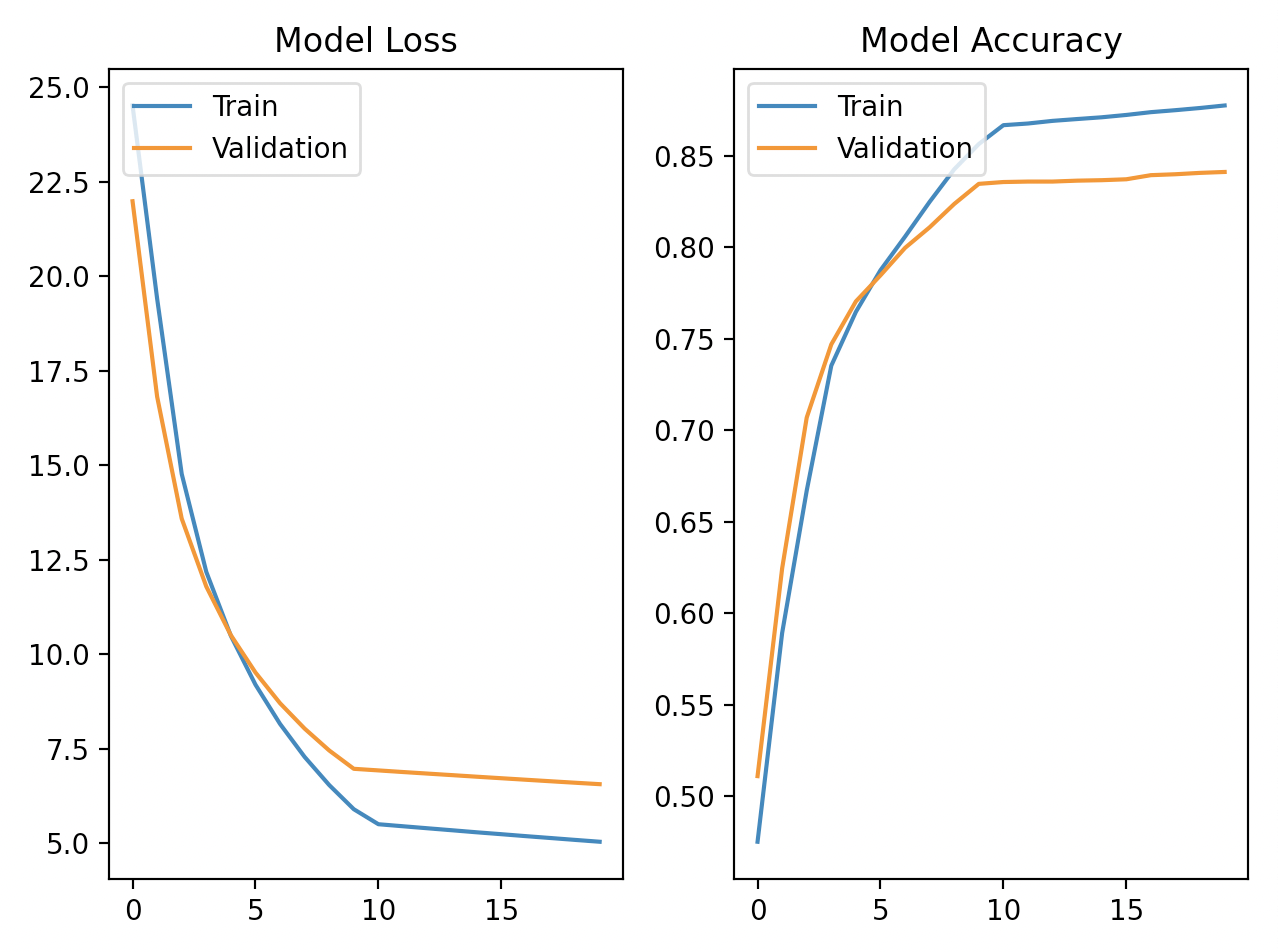}
  \caption{experiment 4}
  \label{fig:ex4} 
\end{figure}

\begin{figure}  
\centering
  \includegraphics[width=100mm,scale=0.5]{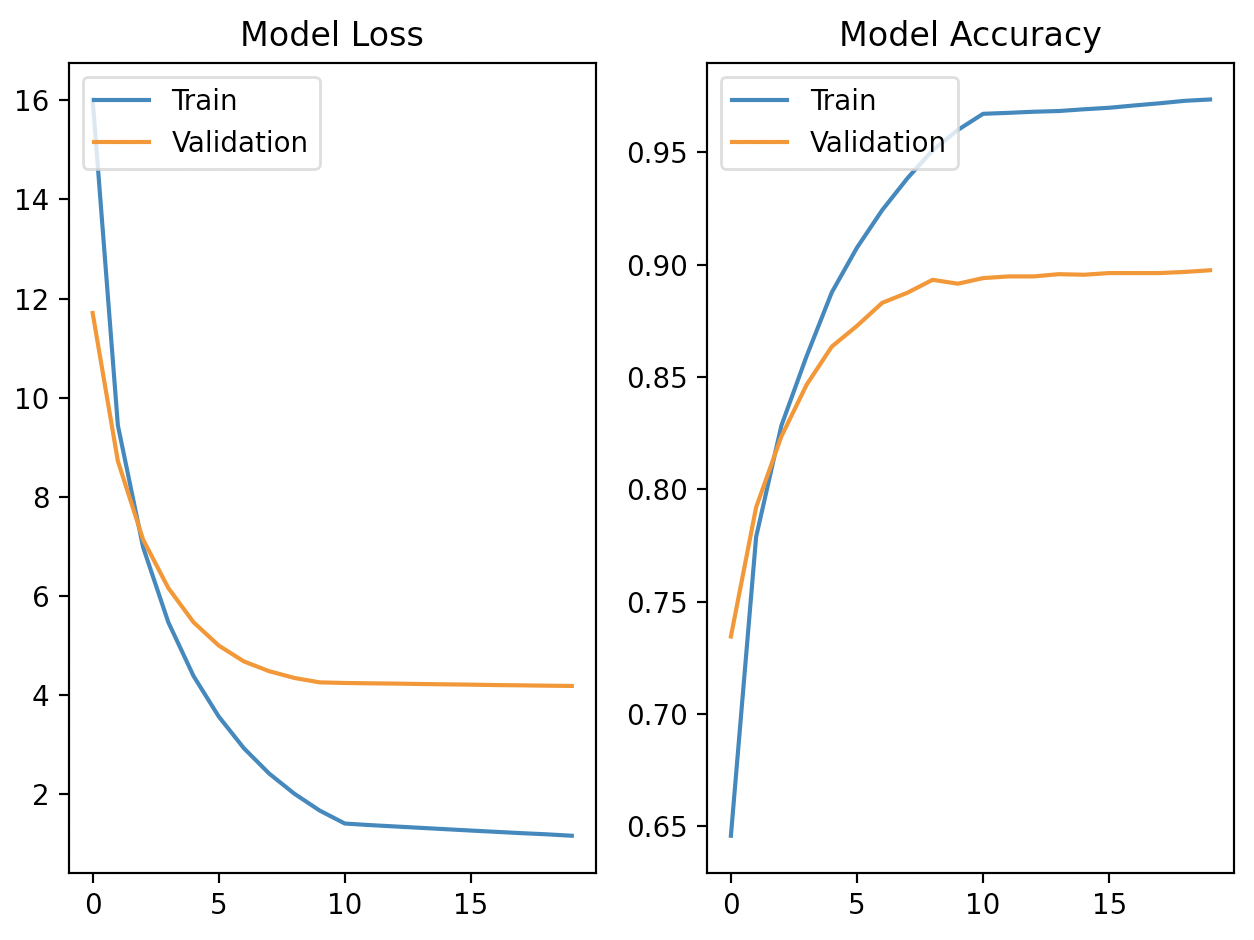}
  \caption{experiment 5}
  \label{fig:ex5} 
\end{figure}

\begin{figure} 
\centering
  \includegraphics[width=100mm,scale=0.5]{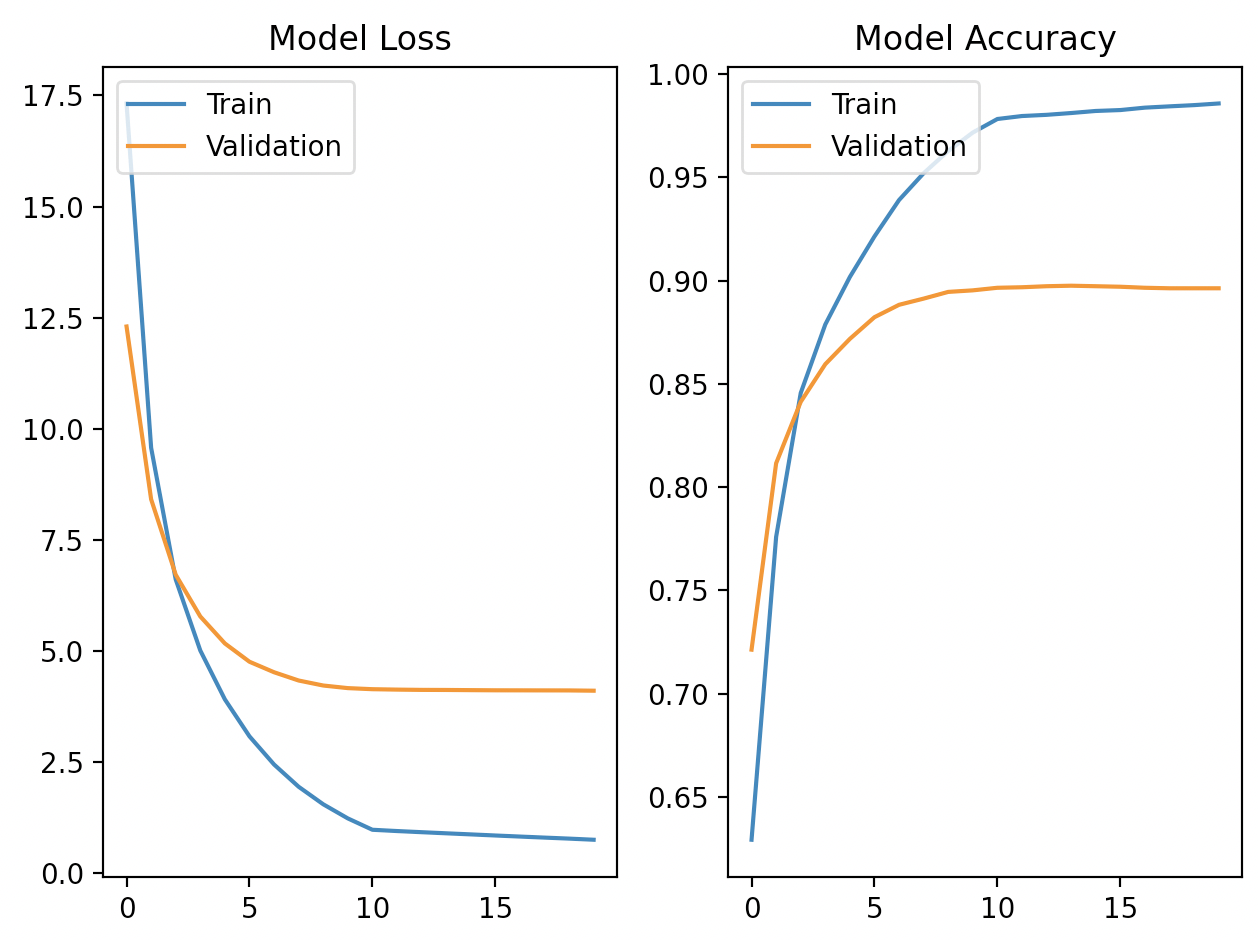}
  \caption{experiment 6}
  \label{fig:ex6} 
\end{figure}

\bibliographystyle{agsm}
\bibliography{semtagging}
\end{document}